\documentclass[pdflatex,sn-mathphys-num]{sn-jnl}

\usepackage{graphicx}%
\usepackage{multirow}%
\usepackage{amsmath,amssymb,amsfonts}%
\usepackage{amsthm}%
\usepackage{mathrsfs}%
\usepackage[title]{appendix}%
\usepackage{xcolor}%
\usepackage{textcomp}%
\usepackage{manyfoot}%
\usepackage{booktabs}%
\usepackage{algorithm}%
\usepackage{algorithmicx}%
\usepackage{algpseudocode}%
\usepackage{placeins}%
\usepackage{listings}
\usepackage{rotating}
\usepackage{makecell}
\usepackage{booktabs}   
\usepackage{tabularx}   
\usepackage{ragged2e}   
\usepackage{caption}    
\usepackage{float}      

\usepackage{tabularx} 
\usepackage{booktabs} 
\usepackage{graphicx}
\usepackage{lscape} 
\usepackage{threeparttable}
\usepackage{tabularx,booktabs,array,ragged2e}
\newcolumntype{P}[1]{>{\RaggedRight\arraybackslash}p{#1}}
\newcolumntype{Y}{>{\RaggedRight\arraybackslash}X}
\newcolumntype{Z}{>{\RaggedRight\arraybackslash\hsize=1.25\hsize}X}

\theoremstyle{thmstyleone}%
\theoremstyle{thmstyletwo}%

\theoremstyle{thmstylethree}%

\begin{document}

\title[A Dual Pipeline Machine Learning Framework for Automated Multi Class Sleep Disorder Screening Using Hybrid Resampling and Ensemble Learning]{A Dual Pipeline Machine Learning Framework for Automated Multi Class Sleep Disorder Screening Using Hybrid Resampling and Ensemble Learning}








\author*[1]{\fnm{Md Sultanul Islam} \sur{Ovi}}\email{movi@gmu.edu}

\author[2]{\fnm{Muhsina Tarannum} \sur{Munfa}}\email{munfa2014@gmail.com}

\author[3]{\fnm{G.M.M Miftahul Alam} \sur{Adib}}\email{2023134010@student.sust.edu}

\author[4]{\fnm{Syed Sabbir} \sur{Hasan}}\email{syed034@student.sust.edu}

\affil*[1]{\orgdiv{Computer Science}, \orgname{George Mason University}, \orgaddress{\street{4400 University Dr}, \city{Fairfax}, \postcode{22030}, \state{Virginia}, \country{USA}}}

\affil[2]{\orgdiv{Computer Science and Engineering}, \orgname{Metropolitan University}, \orgaddress{\city{Sylhet}, \postcode{3104}, \country{Bangladesh}}}

\affil[3]{\orgdiv{Statistics}, \orgname{Shahjalal University of Science and Technology}, \orgaddress{\city{Sylhet}, \postcode{3114}, \country{Bangladesh}}}

\affil[4]{\orgdiv{Industrial and Production Engineering}, \orgname{Shahjalal University of Science and Technology}, \orgaddress{\city{Sylhet}, \postcode{3114}, \country{Bangladesh}}}

\maketitle

\begin{center}
\small
This manuscript is a preprint and is currently under review for journal publication.
\end{center}

\clearpage

\begin{center}
    \textbf{\large Abstract}
\end{center}

Accurate classification of sleep disorders, particularly insomnia and sleep apnea, is important for reducing long term health risks and improving patient quality of life. However, clinical sleep studies are resource intensive and are difficult to scale for population level screening. This paper presents a Dual Pipeline Machine Learning Framework for multi class sleep disorder screening using the Sleep Health and Lifestyle dataset. The framework consists of two parallel processing streams: a statistical pipeline that targets linear separability using Mutual Information and Linear Discriminant Analysis, and a wrapper based pipeline that applies Boruta feature selection with an autoencoder for non linear representation learning. To address class imbalance, we use the hybrid SMOTETomek resampling strategy. In experiments, Extra Trees and K Nearest Neighbors achieved an accuracy of 98.67\%, outperforming recent baselines on the same dataset. Statistical testing using the Wilcoxon Signed Rank Test indicates that the improvement over baseline configurations is significant, and inference latency remains below 400 milliseconds. These results suggest that the proposed dual pipeline design supports accurate and efficient automated screening for non invasive sleep disorder risk stratification.



\section{Introduction}\label{sec1}

Sleep disorders constitute a pervasive global health challenge, severely diminishing cognitive function, physical health, and overall quality of life. Among the most deleterious of these conditions are Insomnia and Obstructive Sleep Apnea (OSA), which increasingly affect the young adult population \cite{ahadian2024adopting}. Insomnia is characterized by chronic difficulties in sleep initiation or maintenance, leading to daytime fatigue and significant cognitive alterations \cite{10462120}. The chronic persistence of these disorders exacerbates the risk of severe comorbidities, including cardiovascular disease and neurocognitive degeneration. Consequently, the accurate and timely diagnosis of emerging sleep pathologies is critical. However, traditional diagnostic modalities, particularly the gold standard polysomnography (PSG), are resource-intensive, necessitate expert invigilation, and impose significant temporal constraints. These limitations render PSG inaccessible for long-duration monitoring and impractical for large-scale population screening \cite{Sarala2025Sleep}.\\

The epidemiological scale of this issue is substantial. Recent reports indicate that OSA alone affects approximately one billion adults aged 30 to 69 worldwide \cite{benjafield2019estimation}. Similarly, insomnia prevalence has surged, affecting nearly 20\% of young adults who experience intermittent symptoms. If left unaddressed, the chronic sleep deficits resulting from these disorders compound cardiometabolic and neurocognitive risks, creating a pressing public health burden \cite{morin2022epidemiology}.\\

To address the scalability limitations of clinical PSG, Machine Learning (ML) has emerged as a promising paradigm for automated disorder screening. Recent advancements in algorithmic efficacy allow for the analysis of complex physiological signals, demographic profiles, and lifestyle factors to predict diagnostic risks with high precision. Ensemble approaches, deep learning neural networks, and hybrid models have demonstrated the capacity to estimate risk accurately \cite{Rahman2025Improving}. Furthermore, the integration of these models with modern wearable sensors facilitates cost-effective, home-based monitoring, thereby mitigating the data acquisition bottlenecks inherent in conventional clinical environments \cite{Monowar2025Advanced, Imtiaz2021A}. ML-based solutions utilizing limited input modalities, such as nocturnal oximetry or photoplethysmography (PPG), have opened new opportunities for risk stratification even when the patient is at home \cite{mashaqi2020combined}. Previous studies employing Convolutional Neural Networks (CNN) and Recurrent Neural Networks (RNN), such as the DeepSleepNet architecture, have shown success in staging from single-channel Electroencephalogram (EEG)\cite{supratak2017deepsleepnet}.\\

Despite these advancements, existing ML-based solutions often struggle with the heterogeneous nature of lifestyle data and severe class imbalance, which can bias predictions toward healthy classes. While recent benchmarks by Rahman et al. \cite{Rahman2025Improving} and Monowar et al. \cite{Monowar2025Advanced} have achieved high accuracy, there remains a need for a framework that rigorously handles nonlinear feature interactions and ensures statistical robustness in its predictions.\\

In this study, we propose a Dual Pipeline Machine Learning Framework (DPMLF) designed to maximize diagnostic precision for multiclass sleep disorder screening. By integrating rigorous statistical feature engineering with advanced wrapper-based selection methods, our approach addresses the limitations of prior works. The primary contributions of this paper are as follows:

\begin{enumerate}
    \item We introduce a novel Dual Pipeline architecture that processes data through two parallel paths: a statistical pipeline focusing on linear separability via Mutual Information (MI) and Linear Discriminant Analysis (LDA), and a wrapper-based pipeline leveraging Boruta and Autoencoders for nonlinear feature extraction.
    \item We implement a hybrid class balancing strategy using SMOTETomek, which simultaneously oversamples minority classes and cleans class boundaries to mitigate bias against Insomnia and Sleep Apnea cases.
    \item We achieve state-of-the-art classification performance using Extra Trees (Extremely Randomized Trees) and K-Nearest Neighbors (KNN) classifiers, outperforming recent 2025 benchmarks on the Sleep Health and Lifestyle dataset.
    \item We provide a rigorous statistical validation using the Wilcoxon Signed-Rank Test to confirm that the performance improvements are statistically significant and not artifacts of random variance.
\end{enumerate}

The remainder of this paper is organized as follows. Section 2 reviews the relevant literature and identifies existing research gaps. Section 3 details the proposed methodology, including data preprocessing, the dual pipeline architecture, and the classification algorithms employed. Section 4 presents the experimental results, ablation studies, and a comprehensive discussion of the findings. Section 5 addresses the ethical considerations inherent in deploying AI for healthcare screening. Finally, Section 6 provides the conclusion and outlines directions for future research.\\

\section{Literature Review}\label{sec2}

The early detection of sleep disorders such as Sleep Apnea and Insomnia is critical for mitigating long-term cardiovascular and neurocognitive risks. While traditional polysomnography remains the gold standard, its resource intensity has driven a paradigm shift toward ML-based screening. This review critically analyzes the evolution of these automated approaches, categorizing them into signal-based and lifestyle-based modalities, to identify the methodological gaps that necessitate the proposed Dual Pipeline Framework.

\subsection{Signal-Based Detection Methodologies}
The majority of existing literature focuses on detecting disorders, particularly Obstructive Sleep Apnea (OSA), using physiological signals like ECG or SpO2. Kristiansen et al. \cite{10.1145/3433987}, and Bahrami et al. \cite{bahrami2022deep} demonstrated that deep learning architectures could achieve accuracies between 85\% and 89\% using multi-signal inputs. Similarly, Serrano Alarcón et al. \cite{alarcon2023obstructive} utilized 1D CNNs on portable monitor signals to detect events with 84.3\% accuracy. While effective, these approaches inherently require physical sensors and continuous monitoring equipment, limiting their scalability for mass population screening or non-contact scenarios. Furthermore, studies like Khandelwal et al. \cite{khandelwal2023ecg} acknowledge that reducing data dimensionality remains a challenge in these high-frequency signal domains. In the context of Insomnia, researchers such as Tripathi et al. \cite{tripathi2022ensemble} and Sharma et al. \cite{sharma2021automatic} successfully leveraged Heart Rate Variability (HRV) features, achieving accuracies exceeding 97\% with ensemble and KNN classifiers. However, these models rely heavily on the availability of precise physiological recordings, which may not be accessible in resource-constrained primary care settings.

\subsection{Lifestyle-Based Screening and Multi-Disorder Classification}
To address the scalability limits of sensor-based systems, recent research has pivoted toward screening based on lifestyle and demographic data. Hidayat et al. \cite{hidayat2023classification} utilized the Sleep Health and Lifestyle dataset to classify multiple disorders, achieving 92\% accuracy with Random Forest. However, their study relied on standard feature selection without deeply addressing the nonlinear interactions between lifestyle factors. Monowar et al. \cite{Monowar2025Advanced} advanced this field by introducing ensemble stacking and SMOTE for class balancing, pushing accuracy to 96.88\%. Despite these improvements, standard ensemble methods often fail to distinctively process linear versus nonlinear feature relationships, potentially capping performance when heterogeneous data types are involved.\\

Other works, such as Ha et al. \cite{ha2023predicting}, employed XGBoost on large cohorts to predict comorbidities but noted that mild OSA remains difficult to screen without expanded phenotypic features. This highlights a recurring limitation in the literature: the struggle to maintain high sensitivity across all classes when dealing with the severe class imbalance typical of medical datasets.

\subsection{Research Gaps and Synthesis}
Despite the proliferation of ML models for sleep disorder diagnosis, significant methodological gaps remain. First, most studies treat feature engineering as a monolithic process, applying the same transformation pipeline to all features regardless of their statistical distribution or linear separability. Second, while class imbalance is acknowledged, few studies implement hybrid resampling techniques (like SMOTETomek) that simultaneously generate synthetic samples and clean decision boundaries. Finally, many high-performing models lack rigorous statistical verification (such as the Wilcoxon Signed-Rank Test) to confirm that performance gains are not merely artifacts of random data splitting.\\

This study addresses these specific gaps by proposing a Dual Pipeline Framework. Unlike prior works that rely on a single processing path, our approach splits features into statistical and wrapper-based pipelines to optimally handle both linear and nonlinear dependencies. By integrating this with hybrid resampling and rigorous statistical validation, we aim to overcome the precision limitations observed in identifying minority disorder classes in previous lifestyle-based studies.

\begin{table*}[t]
\centering
\begin{threeparttable}
\caption{Comprehensive summary of related works analyzing methodologies and identifying research gaps in sleep disorder classification}
\label{tab:related_works}
\scriptsize
\renewcommand{\arraystretch}{1.1}
\setlength{\tabcolsep}{3pt}
{\footnotesize
\begin{tabularx}{\textwidth}{@{}p{1.7cm} p{2.1cm} p{2.3cm} p{1.8cm} X X@{}}
\toprule
\textbf{Author} & \textbf{Used Model} & \textbf{Dataset} & \textbf{Result} & \textbf{Main Contribution} & \textbf{Limitation} \\
\midrule
Cheng et al. (2023) \cite{bib14} &
CNNs (VGG16), Shallow NNs &
PhysioNet CAP Sleep Database (108 participants; multimodal EEG, ECG, EMG) &
Sleep Stage accuracy 94.34\%, F1 0.92; Sleep Disorder accuracy 99.09\%, F1 0.99 &
Distributed multimodal multilabel decision system for sleep stage and disorder recognition using CNN plus shallow NNs &
Single dataset, limited generalization, sleep stage harder than disorder recognition. \\
\midrule
Satapathy et al. (2023) \cite{bib15} &
RF, KNN, SVM, CNN plus LSTM &
Sleep-EDF Dataset   &
CNN+LSTM 87.4\%, KNN 83.65\%, SVM 76.04\%, RF 74.07\% &
Compared ML and DL classifiers for automated sleep stage classification &
Limited to healthy subjects, generalizability unverified, small-scale comparisons. \\
\midrule
Dritsas et al. (2024) \cite{bib17} &
Logistic Regression, SVM (Linear/Polynomial) &
NHANES dataset &
SVM Polynomial accuracy 91.44\%, f-measure 0.914 &
One-vs-All and One-vs-One ML models for sleep disorder prediction &
The dataset is small and lacks detailed clinical information on subjects' profiles \\
\midrule
Kazemi et al. (2025) \cite{bib19} &
1D-Vision Transformer(1D-ViT, multitask explainable architecture) &
Polysomno- graphy(PSG) recordings from UCI Sleep Center (123 subjects) &
Sleep stage classification accuracy 78\%(Cohen’s Kappa ~0.66), Sleep apnea classification 74\%(Cohen’s Kappa ~0.58) &
Multimodal multitask 1D ViT for simultaneous sleep stage and disorder classification with attention interpretability &
Small dataset, focused on respiratory disorders, no real-world deployment. \\
\midrule
Calderon et al. (2024) \cite{bib20} &
Elastic Net, Bayesian Network (CPDAG), SHAP &
31{,}285 U.S. college students (insomnia + seven comorbid disorders) &
$R^2_{\text{train}}$ = 0.44 with RMSE = 5.0, $R^2_{\text{test}}$ = 0.33 with RMSE = 5.47; MDD most predictive &
Combined ML and Bayesian network to explore connections of key disorders with insomnia &
Cross-sectional, self-report bias, limited generalizability, no external replication. \\
\midrule
Jisha et al. (2024) \cite{bib21} &
LSTM, IoT UWB Sensor &
Custom IR UWB dataset &
Accuracy 87–92\%, Sensitivity 87–90\%, Specificity 83–94\% &
Non-contact IoT + LSTM framework for real-time sleep apnea detection &
Small dataset, high LSTM latency, no external validation, limited hardware calibration. \\
\bottomrule
\end{tabularx}
}
\end{threeparttable}
\end{table*}

\FloatBarrier

\section{Methodology}

This study establishes a systematic machine learning framework aimed at maximizing diagnostic precision for multi-class sleep disorders. The proposed methodology adheres to a structured workflow that encompasses data acquisition, rigorous preprocessing, a dual-pipeline approach for feature engineering, hybrid class balancing, and ensemble classification.

\subsection{Dataset Description}

The experimental evaluation utilizes the publicly available Sleep Health and Lifestyle dataset, sourced from Kaggle \cite{tharmalingam_sleep_health_lifestyle_kaggle}. This dataset was specifically selected for its comprehensive integration of physiological parameters with lifestyle determinants, providing a high-fidelity representation of real-world clinical screening scenarios and establishing a robust baseline for algorithmic benchmarking \cite{chung2016stop, buysse1989pittsburgh}. The dataset comprises 374 clinical entries characterized by 13 distinct attributes, including 8 numerical and 5 categorical features, collectively providing a comprehensive profile of human sleep patterns and lifestyle metrics. An operational overview of the dataset characteristics is presented in Table~\ref{tab:dataset_summary}.\\

The feature set encompasses specific demographic, physiological, and lifestyle identifiers, namely \textit{Person ID}, \textit{Gender}, \textit{Age}, \textit{Occupation}, \textit{Sleep Duration}, \textit{Quality of Sleep}, \textit{Physical Activity Level}, \textit{Stress Level}, \textit{BMI Category}, \textit{Blood Pressure}, \textit{Heart Rate}, \textit{Daily Steps}, and the target variable \textit{Sleep Disorder}. A comprehensive description of these attributes, including their respective data types and functional definitions, is detailed in Table~\ref{tab:feature_description}.

\begin{table}[htbp]
\centering
\caption{Operational characteristics and summary of the Sleep Health and Lifestyle dataset.}
\label{tab:dataset_summary}
\begin{tabular}{l p{0.60\linewidth}}
\toprule
\textbf{Description} & \textbf{Details} \\
\midrule
Dataset Name & Sleep Health \& LifeStyle Dataset \\
Source & Kaggle \\
Number of Samples & 374 \\
Number of Features & 13 \\
Target Variable & Sleep Disorder (None, Insomnia, and Sleep Apnea) \\
Missing Values & No \\
Imbalanced Dataset & Yes \\
Data Type & Mixed (Numerical and Categorical) \\
\bottomrule
\end{tabular}
\end{table}

\begin{table}[htbp]
\centering
\caption{Distribution of samples across the three diagnostic classes within the original dataset, highlighting the class imbalance.}
\label{tab:category_samples}
\begin{tabular}{l c}
\toprule
\textbf{Sleep Disorder} & \textbf{Number of Samples} \\
\midrule
None & 219 \\
Sleep Apnea & 78 \\
Insomnia & 77 \\
\bottomrule
\end{tabular}
\end{table}

The target variable, \textit{Sleep Disorder}, classifies subjects into three distinct diagnostic groups: `None' (Healthy), `Sleep Apnea', and `Insomnia'. As illustrated in Table~\ref{tab:category_samples}, the dataset exhibits a significant class imbalance \cite{he2009learning}, with the `None' class constituting the majority (219 samples), while `Sleep Apnea' and `Insomnia' are represented by 78 and 77 samples, respectively. \\

In terms of feature specifics, \textit{Person ID} acts as a unique anonymized identifier. Demographic attributes include \textit{Gender}, \textit{Age}, and \textit{Occupation}. Physiological metrics are represented by \textit{Sleep Duration} (daily hours) and \textit{Heart Rate} (beats per minute). Subjective health assessments include \textit{Quality of Sleep} and \textit{Stress Level}, both quantified on a scale from 1 to 10. Lifestyle factors are captured via \textit{Physical Activity Level} (minutes/day) and \textit{Daily Steps}. The \textit{BMI Category} stratifies subjects into `Underweight', `Normal', and `Overweight' classes. Notably, \textit{Blood Pressure} is recorded as a compound string denoting systolic and diastolic values (mmHg), which necessitates specific preprocessing to isolate the maximum and minimum pressure values during ventricular contraction. A clear understanding of these feature distributions and interactions is critical for effective preprocessing and robust model training.

\begin{table*}[!t]
\centering
\caption{Comprehensive nomenclature and description of the thirteen physiological and lifestyle attributes utilized in the study.}
\label{tab:feature_description}
\footnotesize
\setlength{\tabcolsep}{4pt}
\renewcommand{\arraystretch}{1.25}
\begin{tabularx}{\textwidth}{>{\raggedright\arraybackslash}p{0.22\textwidth} >{\raggedright\arraybackslash}X >{\raggedright\arraybackslash}p{0.12\textwidth}}
\toprule
\textbf{Feature Name} & \textbf{Description} & \textbf{Data Type} \\
\midrule
Person ID & Unique identifier for each individual & Numerical \\
Gender & Gender of each person (Male/Female) & Categorical \\
Age & Age of each person in years & Numerical \\
Occupation & Profession of each person & Categorical \\
Sleep Duration & Number of hours each person sleeps in a day & Numerical \\
Quality of Sleep & Subjective sleep quality score from 1 (poor) to 10 (excellent) & Numerical \\
Physical Activity Level & Daily physical activity duration (minutes/day) & Numerical \\
Stress Level & Subjective stress level score ranging from 1 (low) to 10 (high) & Numerical \\
BMI Category & Body Mass Index (Underweight, Normal, Overweight) & Categorical \\
Blood Pressure & Blood pressure in mmHg, formatted as systolic/diastolic & Categorical \\
Heart Rate (bpm) & Resting heart rate in beats per minute & Numerical \\
Daily Steps & Number of steps taken daily & Numerical \\
Sleep Disorder & Presence of sleep disorder: None, Insomnia, or Sleep Apnea & Categorical \\
\bottomrule
\end{tabularx}
\end{table*}

\begin{figure}[htbp]
    \centering
    \makebox[\linewidth][c]{%
        \includegraphics[width=1.25\linewidth]{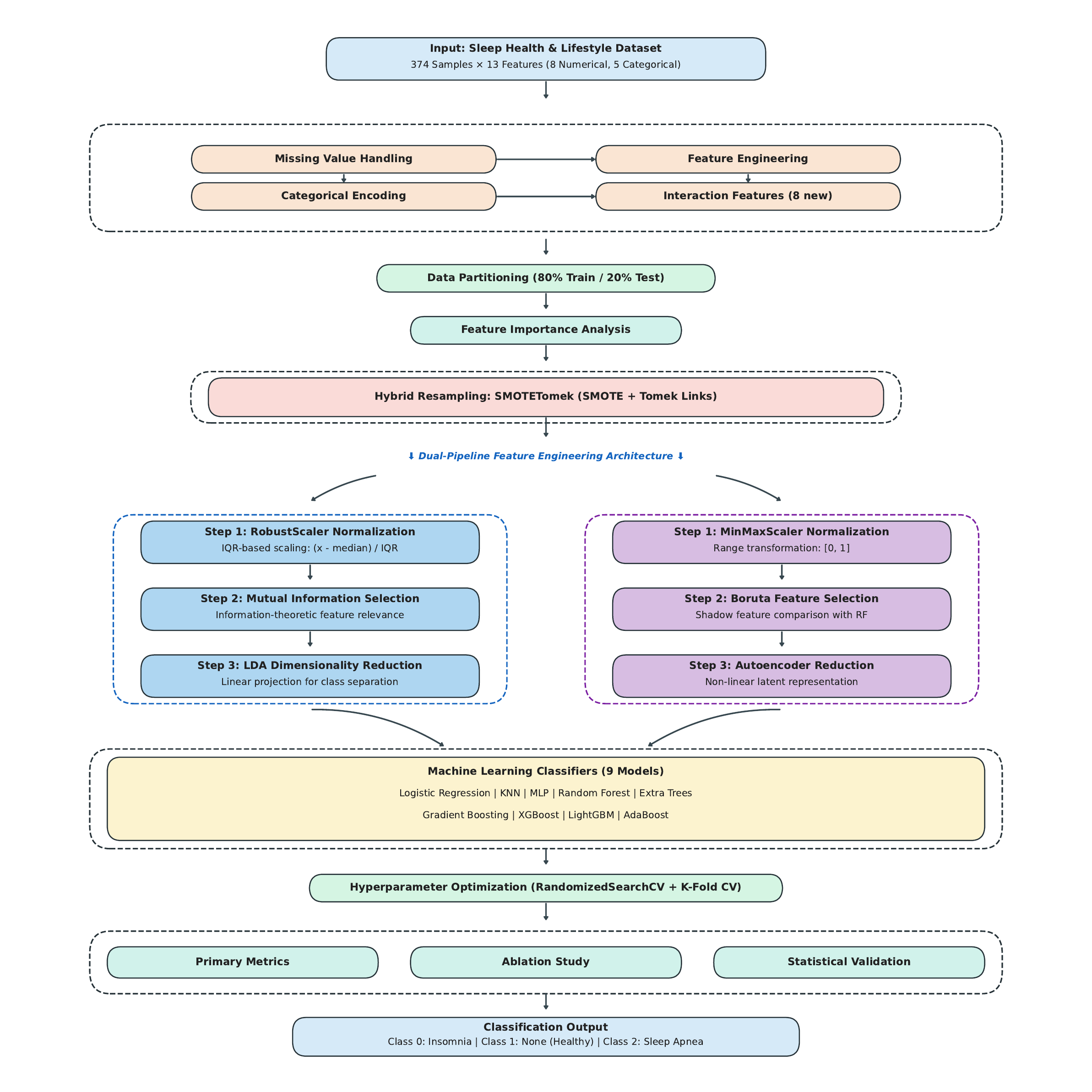}
    }
    \caption{Schematic representation of the proposed Dual-Pipeline Machine Learning Framework, illustrating the parallel processing paths for statistical (Pipeline 1) and wrapper-based (Pipeline 2) feature engineering.}
    \label{fig:methodology_flowchart}
\end{figure}

\subsection{Proposed Framework Workflow}
To address the heterogeneous and imbalanced nature of physiological sleep data, we designed a comprehensive Dual Pipeline Machine Learning Framework. As illustrated in Figure~\ref{fig:methodology_flowchart}, the workflow begins with data acquisition and extensive preprocessing, including categorical encoding and the generation of interaction features to capture non-linear dependencies. A critical component of this architecture is the parallel execution of two distinct feature engineering pipelines: Pipeline 1 (Statistical) applies RobustScaler and Mutual Information, focusing on linear separability via LDA, while Pipeline 2 (Wrapper-based) utilizes MinMaxScaler and Boruta feature selection, leveraging an Autoencoder for nonlinear dimensionality reduction. The processed feature sets converge at the classification stage, where nine supervised algorithms, including advanced ensemble methods, are trained and optimized via Stratified 8-Fold Cross-Validation. This structural design enables a rigorous comparative analysis to identify the optimal preprocessing strategy for minimizing false negatives in sleep disorder screening.

\subsection{Data Preprocessing and Feature Engineering}

Effective data preprocessing is critical for maximizing model generalization and mitigating noise inherent in physiological datasets. We executed a rigorous data preparation pipeline to transform raw clinical entries into a format optimized for ensemble learning.

\begin{itemize}
\item \textbf{Data Cleaning and Partitioning:} A comprehensive data integrity audit confirmed the absence of null values across all features, eliminating the need for imputation. The dataset was stratified and partitioned, reserving 80\% of the samples for training and 20\% for testing to evaluate model performance on unseen data. Feature extraction involved decomposing the compound \textit{Blood Pressure} attribute into distinct \textit{Systolic BP} and \textit{Diastolic BP} features to isolate their individual predictive contributions. Irrelevant identifiers, specifically \textit{Person ID}, were removed to prevent model overfitting to non-informative data. Furthermore, to address sparsity in the \textit{Occupation} feature, infrequent professions were aggregated into an \textit{Other} category, ensuring robust statistical representation \cite{pedregosa2011scikit}.\\

\item \textbf{Feature Encoding:}  Categorical variables underwent specific encoding strategies based on their ordinal or nominal nature. The \textit{BMI Category} was transformed by replacing categorical labels with the mean value of their respective clinical ranges. Nominal variables such as \textit{Occupation} were processed using one-hot encoding. Binary and ordinal variables, including \textit{Gender} and the target variable \textit{Sleep Disorder}, were transformed using label encoding. Specifically, \textit{Gender} was mapped to binary values (\textit{Female}: 0, \textit{Male}: 1), and the target classes were encoded as \textit{Insomnia}: 0, \textit{None}: 1, and \textit{Sleep Apnea}: 2.\\

\item \textbf{Feature Engineering:} To capture nonlinear physiological dependencies that single variables may fail to represent, we engineered a set of interaction features. These derived metrics, detailed in Table~\ref{tab:interaction_features}, quantify relationships such as the ratio of sleep duration to physiological stress markers.
\end{itemize}

\begin{table}[htbp]
\centering
\caption{Mathematical formulation and physiological justification for the engineered interaction features.}
\label{tab:interaction_features}
\small
\setlength{\tabcolsep}{8pt}
\renewcommand{\arraystretch}{1.5}
\begin{tabularx}{\linewidth}{>{\raggedright\arraybackslash}p{0.26\linewidth} @{\hspace{20pt}} >{\raggedright\arraybackslash}p{0.28\linewidth} >{\raggedright\arraybackslash}X}
\toprule
\textbf{Feature Name} & \textbf{Formula} & \textbf{Description} \\
\midrule
\textit{Stress\_sleep\_interaction} & $\frac{\text{Stress Level}}{\text{Quality of Sleep}}$ & Quantifies the inverse correlation between perceived stress and subjective sleep quality \\
\textit{Sleep\_Heart\_ratio} & $\frac{\text{Sleep Duration}}{\text{Heart Rate}}$ & Reflects physiological rest efficiency relative to cardiovascular load \\
\textit{Sleep\_Steps\_Ratio} & $\frac{\text{Sleep Duration}}{\text{Daily Steps}}$ & Indicates the balance between daily physical exertion and recovery time \\
\textit{Sleep\_Stress\_Ratio} & $\frac{\text{Sleep Duration}}{\text{Stress Level}}$ & Measures the adequacy of sleep relative to reported stress levels \\
\textit{BMI\_Activity} & $\text{BMI} \times \text{Activity Level}$ & Captures the combined influence of body composition and physical exertion \\
\textit{Pulse\_Pressure} & $\text{Systolic} - \text{Diastolic}$ & Represents arterial stiffness and cardiovascular risk \\
\textit{log\_steps} & $\log(\text{Daily Steps})$ & Normalizes the distribution of step counts to reduce skewness \\
\textit{sqrt\_sleep} & $\sqrt{\text{Sleep Duration}}$ & Stabilizes variance and mitigates the impact of outliers in sleep data \\
\bottomrule
\end{tabularx}
\end{table}

\begin{figure}[htbp]
    \centering
    \includegraphics[width=.95\linewidth]{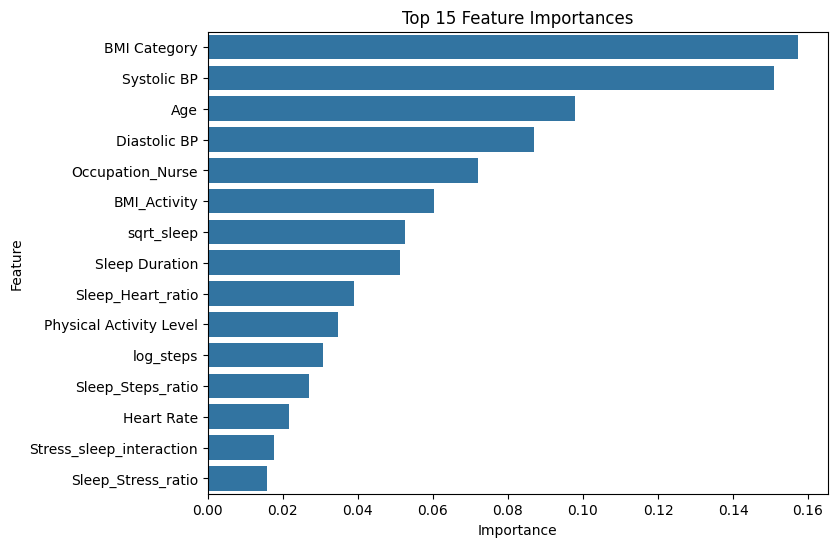}
    \caption{Feature Importance Analysis illustrating the relative predictive power of the original and engineered features derived from tree-based ensemble estimators.}
    \label{fig:feature_importance}
\end{figure}

Following feature engineering, we conducted a feature importance analysis using tree-based estimators to rank predictors by their contribution to information gain \cite{breiman2001random, geurts2006extremely}. This analysis, visualized in Figure~\ref{fig:feature_importance}, guided the subsequent selection processes in our dual pipeline framework.

\subsection{Handling Class Imbalance with SMOTETomek}

\begin{table}[htbp]
\centering
\caption{Comparison of class distribution frequencies before and after applying the SMOTETomek hybrid resampling technique.}
\label{tab:resampling}
\begin{tabular}{l c c}
\toprule
\textbf{Classes} & \textbf{Original Count} & \textbf{Resampled Count} \\
\midrule
Insomnia (0) & 62 & 175 \\
None (1) & 175 & 173 \\
Sleep Apnea (2) & 62 & 171 \\
\bottomrule
\end{tabular}
\end{table}

\begin{figure}[htbp]
    \centering
    \includegraphics[width=.8\linewidth]{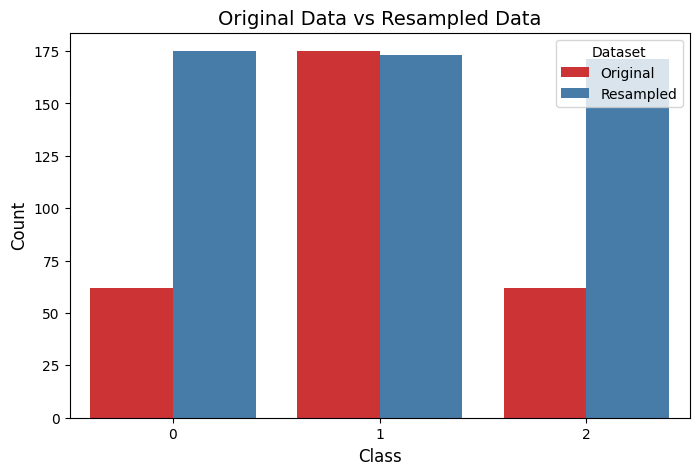}
    \caption{Visualization of the feature space distribution illustrating the transition from the imbalanced original dataset to the balanced dataset following SMOTETomek application.}
    \label{fig:resampling}
\end{figure}

The initial exploratory data analysis revealed a substantial class imbalance within the training set. The majority class (\textit{None}) comprised 175 samples, significantly outnumbering the minority classes, \textit{Insomnia} and \textit{Sleep Apnea}, which contained only 62 samples each. Such disparities often lead to classifiers biased toward the majority class, resulting in poor sensitivity for detecting disorders.

To mitigate this, we employed \textbf{SMOTETomek}, a hybrid resampling technique that synergizes oversampling and undersampling \cite{lemaavztre2017imbalanced, chawla2002smote, batista2004study}. First, the Synthetic Minority Oversampling Technique (SMOTE) generates synthetic examples for the minority classes (\textit{Insomnia} and \textit{Sleep Apnea}) by interpolating between existing samples in the feature space. Subsequently, Tomek Links are identified and removed. A Tomek Link exists if two samples from different classes are each other's nearest neighbors; removing these ambiguous samples cleans the decision boundaries and improves class separability \cite{he2009learning}.

Table~\ref{tab:resampling} details the quantitative impact of this procedure, showing the equalization of class distributions. The visual impact of this resampling on the feature space is demonstrated in Figure~\ref{fig:resampling}, highlighting the enhanced density of minority classes and the clearer demarcation of decision regions.

\subsection{Dual-Pipeline Feature Engineering Architecture}

We implemented a dual-pipeline framework to systematically compare traditional statistical transformations against advanced nonlinear engineering techniques \cite{pedregosa2011scikit}. This architecture enables the extraction of complementary physiological information through distinct normalization, feature selection, and dimensionality reduction strategies.

\subsubsection{Pipeline 1: Statistical Engineering (Linear Focus)}

\begin{figure}[htbp]
    \centering
    \makebox[\linewidth][c]{%
        \includegraphics[width=1.2\linewidth]{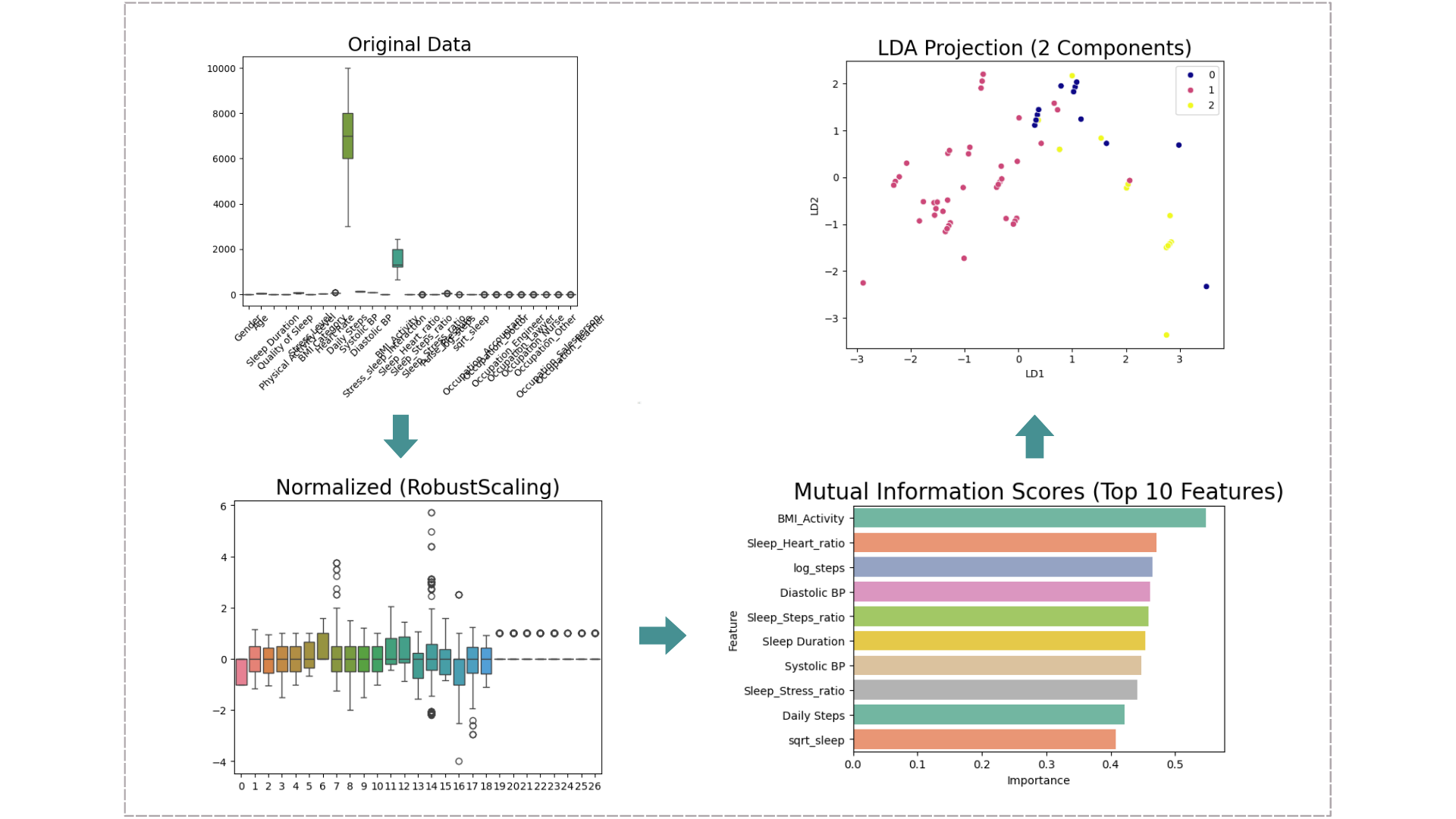}
    }
    \caption{Schematic overview of Pipeline 1 showing the sequential application of outlier-robust scaling, mutual information-based selection, and linear projection via LDA to optimize the feature space for linear separability.}
    \label{fig:pipeline1}
\end{figure}

The first pipeline prioritizes statistical robustness and linear separability, targeting models that assume Gaussian distributions. The complete workflow is illustrated in Figure~\ref{fig:pipeline1}.

\begin{itemize}
    \item \textbf{Normalization:} We employ \textit{RobustScaler} to normalize the feature set using the interquartile range (IQR). This method effectively mitigates the impact of physiological outliers in measurements like heart rate and blood pressure, preventing extreme values from distorting the learning process \cite{rousseeuw2003robust}.
    \item \textbf{Feature Selection:} Feature relevance is assessed using \textit{Mutual Information (MI)}. This information-theoretic method quantifies the dependency between each feature and the target variable, isolating predictors that contribute significant information gain \cite{cover2006elements, battiti1994using}.
    \item \textbf{Dimensionality Reduction:} The final stage employs \textit{Linear Discriminant Analysis (LDA)}, which projects the feature space into a lower-dimensional subspace maximizing the ratio of between-class to within-class variance, thereby optimizing class separability.
\end{itemize}

\subsubsection{Pipeline 2: Wrapper-based Engineering (Non-Linear Focus)}

\begin{figure}[htbp]
    \centering
    \makebox[\linewidth][c]{%
        \includegraphics[width=1.2\linewidth]{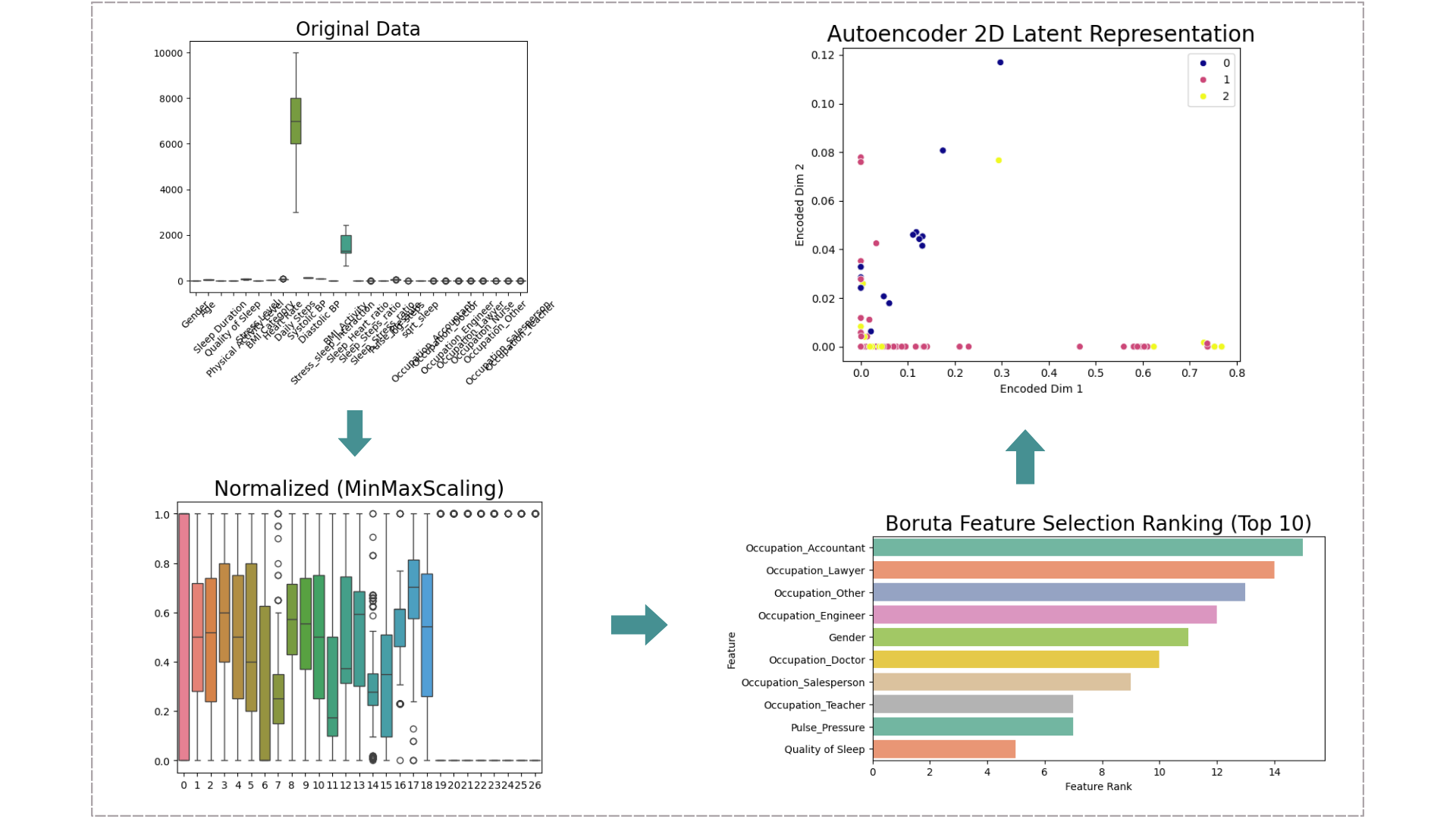}
    }
    \caption{Schematic overview of Pipeline 2 illustrating the workflow of range-based normalization, all-relevant feature selection using Boruta, and non-linear dimensionality reduction via an Autoencoder network.}
    \label{fig:pipeline2}
\end{figure}

The second pipeline is architected to capture complex, non-linear relationships via wrapper methods and neural network compression. Figure~\ref{fig:pipeline2} depicts the sequential stages of this pipeline.

\begin{itemize}
    \item \textbf{Normalization:} We utilize \textit{MinMaxScaler} to transform features into a bounded [0, 1] range. This normalization preserves the original distribution shape while ensuring efficient convergence for gradient-based optimization algorithms.
    \item \textbf{Feature Selection:} We implement the \textit{Boruta} algorithm, a wrapper method built around the Random Forest classifier. Boruta iteratively compares original attributes against randomized \textit{shadow features} to confirm statistical significance, ensuring the model relies on all-relevant predictors rather than merely non-redundant ones \cite{kursa2010feature, breiman2001random}.
    \item \textbf{Dimensionality Reduction:} Dimensionality reduction is achieved via an \textit{Autoencoder}. This unsupervised neural network learns a compressed latent representation of the input data. By minimizing reconstruction error, the Autoencoder captures nonlinear latent structures that linear methods may miss \cite{hinton2006reducing, goodfellow2016deep}.
\end{itemize}

\subsection{Machine Learning Classifiers}
To ensure a rigorous evaluation of the proposed dual pipeline framework, we employed a diverse suite of nine supervised learning algorithms, categorized into linear, non-linear, and ensemble paradigms. This selection allows for a comprehensive analysis of how different inductive biases interact with the engineered feature sets. \\

\textbf{Linear and Non-Linear Estimators:} We utilized Logistic Regression as a baseline linear classifier \cite{mccullagh2019generalized} due to its interpretability and efficacy in high-dimensional spaces where linear separability is assumed. To capture local manifold structures within the physiological data, we employed K-Nearest Neighbors (KNN) \cite{cover1967nearest}, a non-parametric method that classifies instances based on feature similarity. Furthermore, to address complex non-linear dependencies without manual feature interaction, we implemented a Multi-Layer Perceptron (MLP) \cite{goodfellow2016deep, rumelhart1986learning}. This neural network architecture leverages hidden layers and nonlinear activation functions to approximate complex decision boundaries that linear models may fail to resolve. \\

\textbf{Tree-Based Ensemble Methods:} Given the tabular nature of the dataset, we prioritized tree-based ensemble methods for their robustness against overfitting and variance. We implemented Random Forest \cite{breiman2001random}, which employs bagging to construct a multitude of uncorrelated decision trees, determining the final classification via majority voting. To further mitigate variance, we utilized Extra Trees (Extremely Randomized Trees) \cite{geurts2006extremely}, which introduces additional stochasticity by selecting random cut-points for node splitting rather than optimizing for the best split. The boosting paradigm was represented by Gradient Boosting, XGBoost, LightGBM, and AdaBoost \cite{friedman2001greedy, chen2016xgboost, ke2017lightgbm, freund1997decision}. Gradient Boosting optimizes a differentiable loss function by sequentially adding weak learners to correct predecessor errors. XGBoost and LightGBM were selected for their optimized system design and scalability; XGBoost employs a sparsity-aware algorithm and weighted quantile sketch, while LightGBM utilizes gradient-based one-sided sampling (GOSS) for faster training speed. AdaBoost was included to evaluate the efficacy of adaptive re-weighting, where misclassified instances gain stronger influence in subsequent iterations, thereby forcing the model to focus on hard-to-classify samples.\\

\textbf{Hyperparameter Optimization Strategy:} Both \textit{GridSearchCV} and \textit{RandomizedSearchCV} were explored during hyperparameter tuning. \textit{RandomizedSearchCV} allows efficient sampling of large hyperparameter spaces with lower computational cost, while \textit{GridSearchCV} evaluates all possible parameter combinations, which can be computationally expensive and time-consuming. Therefore, to optimize model generalization, we utilized \textit{RandomizedSearchCV} rather than an exhaustive grid search \cite{bergstra2012random}. This approach samples a fixed number of parameter settings from specified distributions, offering a superior trade-off between computational efficiency and optimization performance in high-dimensional spaces. The validation process was conducted using Stratified 8-Fold Cross-Validation \cite{kohavi1995study}. While 5-fold or 10-fold schemes are standard, we selected $k=8$ based on the dataset size ($N=374$). This configuration assigns approximately 47 samples to each validation fold. This split provides a statistically significant representation of minority classes (Sleep Apnea and Insomnia) in the validation set, which might be diluted in a 10-fold split, while retaining a larger portion of data for training compared to a 5-fold split, thus ensuring the model learns robust patterns from the limited data available.

\subsection{Performance Evaluation Strategy}

To objectively assess the diagnostic capability of the proposed framework, we employed a multi-faceted evaluation strategy focusing on classification metrics, statistical significance, and computational efficiency.

\subsubsection{Classification Metrics}
Given the clinical context and class imbalance, relying solely on accuracy can be misleading \cite{he2009learning, sokolova2009systematic, powers2020evaluation}. Therefore, we evaluated performance using four core metrics: Accuracy, Precision, Recall (Sensitivity), and the F1-Score.

\textbf{Accuracy} measures the overall correctness of the model across all classes:
\begin{equation}
\text{Accuracy} = \frac{TP + TN}{TP + TN + FP + FN}
\end{equation}

\textbf{Precision} quantifies the reliability of positive predictions, indicating the proportion of identified sleep disorders that were correct:
\begin{equation}
\text{Precision} = \frac{TP}{TP + FP}
\end{equation}

\textbf{Recall} is critical in medical screening as it measures the model's ability to detect all positive instances of a disorder:
\begin{equation}
\text{Recall} = \frac{TP}{TP + FN}
\end{equation}

\textbf{F1-Score} provides the harmonic mean of Precision and Recall, offering a single metric that balances both concerns, particularly useful for imbalanced distributions:
\begin{equation}
\text{F1-Score} = 2 \times \frac{\text{Precision} \times \text{Recall}}{\text{Precision} + \text{Recall}}
\end{equation}

Where $TP$, $TN$, $FP$, and $FN$ represent True Positives, True Negatives, False Positives, and False Negatives, respectively.

\subsubsection{Computational Efficiency Analysis}
In clinical environments, the latency of a diagnostic system is as critical as its accuracy. To analyze the operational cost of the dual-pipeline framework, we recorded the computational time for two distinct phases:
\begin{enumerate}
    \item \textbf{Training Time ($T_{train}$):} The duration required for the model to learn parameters from the training data, reflecting the computational burden of the pipeline's feature engineering and optimization processes.
    \item \textbf{Testing Time ($T_{test}$):} The latency observed when generating predictions for unseen data.
\end{enumerate}
This analysis allows for an evaluation of the trade-off between algorithmic complexity and real-time diagnostic utility.

\subsubsection{Statistical Significance Testing}
To demonstrate that the performance improvements of our proposed framework are statistically significant and not merely artifacts of random data splitting, we employed the \textbf{Wilcoxon Signed-Rank Test} \cite{wilcoxon1945individual}. This non-parametric statistical hypothesis test is preferred over the Paired t-test for comparing machine learning classifiers because classifier accuracy distributions on different folds often violate the assumption of normality \cite{demvsar2006statistical}.

The Wilcoxon test compares the differences between paired measurements (e.g., the accuracy of the proposed model vs. the baseline on the same cross-validation fold). The test statistic $W$ is calculated as the minimum of the sums of the signed ranks:
\begin{equation}
W = \min(R+, R-)
\end{equation}
where $R+$ is the sum of ranks for positive differences and $R-$ is the sum of ranks for negative differences. A $p$-value less than 0.05 indicates a statistically significant difference between the models, confirming the superiority of the proposed approach.

\section{Results and Discussion}

The empirical evaluation of the proposed framework is divided into two primary phases. The first phase establishes a baseline by assessing the performance of machine learning models on the raw dataset. The second phase analyzes the efficacy of the Dual Pipeline Framework, highlighting the performance gains achieved through specific combinations of feature engineering and hybrid resampling. A detailed decomposition of these components, including the impact of SMOTETomek and feature selection dynamics, is reserved for the subsequent ablation study.

\subsection{Baseline Model Performance}

The initial experimental phase involved training nine distinct machine learning classifiers on the original dataset without applying specific feature engineering or class balancing techniques. Table~\ref{tab6} presents the baseline quantitative metrics, including Accuracy, F1 Score, Recall, and Precision.

\begin{table}[htbp]
\centering
\caption{Baseline classification metrics evaluated on the raw unprocessed dataset across nine distinct classifiers}
\label{tab6}
\begin{tabular*}{\textwidth}{@{\extracolsep{\fill}}ll c c c c}
\toprule
ML Model & Accuracy & F1 Score & Recall & Precision\\
\midrule
Logistic Regression & 97.333\% & 95.658\% & 95.556\% & 96.296\% \\
K-Nearest Neighbors & 92.000\% & 89.526\% & 91.200\% & 88.194\%\\
Random Forest & 94.667\% & 91.998\% & 92.576\% & 92.173\% \\
XGBoost & 96.000\% & 93.521\% & 93.472\% & 93.698\% \\
Gradient Boosting & 97.333\% & 96.328\% & 95.833\% & 97.176\%\\
Extra Trees & 96.000\% & 93.521\% & 93.472\% & 93.698\% \\
AdaBoost & 93.333\% & 89.247\% & 89.306\% & 89.306\% \\
MLP Classifier & 61.333\% & 32.479\% & 37.500\% & 53.425\%\\
LightGBM & 97.333\% & 95.694\% & 95.694\% & 95.694\% \\
\bottomrule
\end{tabular*}
\end{table}

The results indicate that ensemble and linear models possess inherent predictive capabilities even in the absence of optimization. Specifically, Logistic Regression, Gradient Boosting, and LightGBM demonstrated superior baseline performance, each achieving an accuracy of 97.333\%. This suggests that a significant portion of the variance in the dataset is linearly separable. Conversely, the Multi-Layer Perceptron (MLP) Classifier exhibited poor performance with an Accuracy of 61.333\% and an F1 Score of 32.479\%. This suboptimal result for the MLP is attributed to the lack of feature normalization in the raw data, which disproportionately affects neural network convergence. These baseline metrics serve as a critical benchmark for evaluating the relative improvements offered by the proposed dual pipeline architecture.

\subsection{Performance of Optimized Dual-Pipeline Framework}

\begin{table}[htbp]
\centering
\caption{Optimal performance metrics for Pipeline 1 utilizing statistical feature engineering and hybrid resampling strategies}
\label{tab:pipeline1_robust_mi_lda}
\setlength{\tabcolsep}{4pt}
\renewcommand{\arraystretch}{1.15}
\begin{tabular*}{\textwidth}{@{\extracolsep{\fill}}ll c c c c}
\toprule
ML Model & Configuration & Accuracy & F1 Score & Recall & Precision\\
\midrule
Logistic Regression & RobustScaler + SMOTETomek & 96.000\% & 94.379\% & 94.798\% & 94.737\% \\
K-Nearest Neighbors & MI + SMOTETomek & 98.667\% & 97.850\% & 97.917\% & 97.917\% \\
Random Forest & LDA & 96.000\% & 93.521\% & 93.472\% & 93.698\%\\
XGBoost Model & MI + SMOTETomek & 98.667\% & 97.850\% & 97.917\% & 97.917\%\\
Gradient Boosting & RobustScaler & 96.000\% & 94.177\% & 93.611\% & 94.815\%\\
Extra Trees & RobustScaler & 97.333\% & 95.658\% & 95.556\% & 96.296\%\\
AdaBoost & MI & 97.333\% & 95.694\% & 95.694\% & 95.694\% \\
MLP Classifier & MI & 96.000\% & 93.548\% & 93.611\% & 93.611\% \\
LightGBM & MI + SMOTETomek & 98.667\% & 97.850\% & 97.917\% & 97.917\% \\
\bottomrule
\end{tabular*}
\end{table}

\begin{table}[htbp]
\centering
\caption{Optimal performance metrics for Pipeline 2 utilizing wrapper-based feature selection and non-linear dimensionality reduction}
\label{tab:pipeline2_boruta_autoencoder}
\setlength{\tabcolsep}{4pt}
\renewcommand{\arraystretch}{1.15}
\begin{tabular*}{\textwidth}{@{\extracolsep{\fill}}ll c c c c}
\toprule
ML Model & Configuration & Accuracy & F1 Score & Recall & Precision\\
\midrule
Logistic Regression & MinMaxScaler & 97.333\% & 95.658\% & 95.556\% & 96.296\% \\
K-Nearest Neighbors & Autoencoder & 93.333\% & 92.324\% & 93.422\% & 91.799\% \\
Random Forest & MinMaxScaler + SMOTETomek & 97.333\% & 95.658\% & 95.556\% & 96.296\%\\
XGBoost Model & MinMaxScaler + SMOTETomek & 97.333\% & 95.694\% & 95.694\% & 95.694\%\\
Gradient Boosting & Boruta & 96.000\% & 93.521\% & 93.472\% & 93.698\%\\
Extra Trees & Boruta + SMOTETomek & 98.667\% & 97.840\% & 97.778\% & 98.039\%\\
AdaBoost & Autoencoder & 94.667\% & 93.780\% & 94.179\% & 93.669\% \\
MLP Classifier & Autoencoder + SMOTETomek & 93.333\% & 91.676\% & 93.422\% & 90.278\% \\
LightGBM & MinMaxScaler & 97.333\% & 95.694\% & 95.694\% & 95.694\% \\
\bottomrule
\end{tabular*}
\end{table}

The core contribution of this study lies in the application of a dual-pipeline framework designed to maximize diagnostic accuracy through targeted preprocessing. Table~\ref{tab:pipeline1_robust_mi_lda} details the optimal configurations and resulting metrics for Pipeline 1, which focuses on statistical feature engineering using RobustScaler, Mutual Information (MI), and Linear Discriminant Analysis (LDA).

The implementation of Pipeline 1 yielded substantial performance improvements over the baseline. Notably, the K-Nearest Neighbors (KNN) classifier, when coupled with Mutual Information feature selection and SMOTETomek resampling, achieved a peak accuracy of 98.667\% with a corresponding F1 Score of 97.850\%. This represents a significant enhancement over its baseline accuracy of 92.000\%, indicating that the removal of noisy features and the mitigation of class imbalance are critical for distance-based classifiers. Similarly, XGBoost and LightGBM reached the 98.667\% accuracy threshold under the same configuration. The results in Table~\ref{tab:pipeline1_robust_mi_lda} demonstrate that the correct pairing of statistical feature selection and hybrid resampling significantly improves inter-class separability and mitigates the bias toward the majority class.

Parallel to the statistical approach, Table~\ref{tab:pipeline2_boruta_autoencoder} presents the results from Pipeline 2, which leverages wrapper-based feature selection (Boruta) and non-linear dimensionality reduction (Autoencoder) alongside MinMaxScaler. The Extra Trees classifier, integrated with Boruta feature selection and SMOTETomek, achieved the global best performance for this pipeline with an accuracy of 98.667\% and a Precision of 98.039\%. This result underscores the robustness of randomized decision trees when provided with an all-relevant feature subset identified by Boruta. Furthermore, Logistic Regression and Random Forest maintained stable high performance metrics exceeding 97\% within this pipeline. While the Autoencoder configuration did not outperform the tree-based ensembles, it provided competitive results for the MLP classifier, suggesting that non-linear compression remains a viable strategy for specific neural architectures.

The comparative analysis of Table~\ref{tab:pipeline1_robust_mi_lda} and Table~\ref{tab:pipeline2_boruta_autoencoder} reveals that while tree-based ensemble methods remain largely robust across different preprocessing strategies, distance-based and neural models exhibit high sensitivity to the choice of pipeline. To systematically understand the drivers behind these performance shifts, specifically the impact of normalization, resampling, and feature selection dynamics, a comprehensive ablation study and component analysis are presented in the following sections.

\subsection{Ablation Study and Component Analysis}

To isolate the specific contribution of each preprocessing module within the proposed dual pipeline framework, we conducted a comprehensive ablation study. This analysis systematically evaluates how the incremental addition of normalization, class balancing, feature selection, and dimensionality reduction influences model performance. Table~\ref{tab:ablation_pipeline1} and Table~\ref{tab:ablation_pipeline2} present the performance trajectories for the representative models of Pipeline 1 (K-Nearest Neighbors) and Pipeline 2 (Extra Trees), respectively.

\begin{table}[htbp]
\centering
\caption{Stepwise ablation study of the K Nearest Neighbors classifier within Pipeline 1, evaluating the impact of statistical feature engineering}
\label{tab:ablation_pipeline1}
\setlength{\tabcolsep}{4pt}
\renewcommand{\arraystretch}{1.2}
\begin{tabular}{|l|c|c|c|c|c|c|}
\hline
\textbf{Configuration} & \textbf{Resampling} & \textbf{Normalization} & \makecell{\textbf{Feature}\\\textbf{Selection}} & \makecell{\textbf{Dim}\\\textbf{Reduction}} & \textbf{Accuracy} & \textbf{F1 Score}\\
\hline
Baseline & - & - & - & - & 92.000\% & 89.526\%\\
\hline
RobustScaler & - & \checkmark & - & - & 94.667\% & 92.142\%\\
\hline
\makecell{SMOTETomek \\+ RobustScaler} & \checkmark & \checkmark & - & - & 89.333\% & 85.604\%\\
\hline
\makecell{RobustScaler \\+ MI} & - & \checkmark & \checkmark & - & 96.000\% & 94.308\%\\
\hline
\makecell{SMOTETomek \\+ RobustScaler \\+ MI} & \checkmark & \checkmark & \checkmark & - & 98.667\% & 97.850\%\\
\hline
\makecell{RobustScaler \\+ MI + LDA} & - & \checkmark & \checkmark & \checkmark & 96.000\% & 93.521\%\\
\hline
\makecell{SMOTETomek \\+ RobustScaler \\+ MI + LDA} & \checkmark & \checkmark & \checkmark & \checkmark & 96.000\% & 93.521\%\\
\hline
\end{tabular}
\end{table}

\begin{table}[htbp]
\centering
\caption{Stepwise ablation study of the Extra Trees classifier within Pipeline 2, evaluating the impact of wrapper-based selection and autoencoders}
\label{tab:ablation_pipeline2}
\setlength{\tabcolsep}{4pt}
\renewcommand{\arraystretch}{1.2}
\begin{tabular}{|l|c|c|c|c|c|c|}
\hline
\textbf{Configuration} & \textbf{Resampling} & \textbf{Normalization} & \makecell{\textbf{Feature}\\\textbf{Selection}} & \makecell{\textbf{Dim}\\\textbf{Reduction}} & \textbf{Accuracy} & \textbf{F1 Score}\\
\hline
Baseline & - & - & - & - & 96.000\% & 93.521\%\\
\hline
MinMaxScaler & - & \checkmark & - & - & 96.000\% & 93.439\%\\
\hline
\makecell{SMOTETomek \\+ MinMaxScaler} & \checkmark & \checkmark & - & - & 94.667\% & 92.210\%\\
\hline
\makecell{MinMaxScaler \\+ Boruta} & - & \checkmark & \checkmark & - & 96.000\% & 93.521\%\\
\hline
\makecell{SMOTETomek \\+ MinMaxScaler \\+ Boruta} & \checkmark & \checkmark & \checkmark & - & 98.667\% & 97.840\%\\
\hline
\makecell{MinMaxScaler \\+ Boruta \\+ Autoencoder} & - & \checkmark & \checkmark & \checkmark & 93.333\% & 92.285\%\\
\hline
\makecell{SMOTETomek \\+ MinMaxScaler \\+ Boruta \\+ Autoencoder} & \checkmark & \checkmark & \checkmark & \checkmark & 92.000\% & 91.053\%\\
\hline
\end{tabular}
\end{table}

\subsubsection{Efficacy of Hybrid Resampling}
The SMOTETomek hybrid resampling technique was employed to enhance recall rates for the minority classes (Insomnia and Sleep Apnea) by synthetically generating minority samples while simultaneously cleaning class boundaries. The impact of this module varied significantly across model architectures. For distance-based algorithms like K Nearest Neighbors and boosting frameworks like LightGBM, the integration of SMOTETomek yielded substantial gains. Specifically, Table~\ref{tab:ablation_pipeline1} shows that combining SMOTETomek with Mutual Information feature selection elevated the K Nearest Neighbors accuracy from 96.0\% to 98.667\% and the F1 Score to 97.850\%. Conversely, linear models such as Logistic Regression did not benefit from this oversampling strategy, observing a slight degradation in performance metrics. This suggests that while resampling is critical for defining local decision boundaries in non-parametric models, it provides limited utility for models relying on global linear separation.

\subsubsection{Impact of Data Normalization}
The necessity of feature scaling proved to be highly model-dependent. The Multi-Layer Perceptron classifier exhibited a critical dependency on normalization, improving from a baseline accuracy of 61.333\% to 96.000\% after the application of RobustScaler. Similarly, the K Nearest Neighbors model improved from 92.0\% to 94.667\% accuracy upon scaling, as the algorithm requires uniform feature magnitude to calculate Euclidean distances effectively. In contrast, tree-based ensembles such as Random Forest and Extra Trees demonstrated inherent robustness to feature scaling, maintaining consistent accuracy regardless of the normalization technique applied (RobustScaler or MinMaxScaler). These findings confirm that while normalization is a mandatory preprocessing step for neural and distance-based classifiers, it is redundant for decision tree ensembles.

\subsubsection{Feature Selection Dynamics: Mutual Information vs. Boruta}
We compared the efficacy of filter-based Mutual Information against the wrapper-based Boruta algorithm. Mutual Information proved particularly effective for models sensitive to noise, such as K-Nearest Neighbors and AdaBoost. As indicated in Table~\ref{tab:ablation_pipeline1}, the inclusion of Mutual Information features enabled the K-Nearest Neighbors model to achieve its peak performance by filtering out redundant dimensions that distort distance calculations. Although the Extra Trees model in Pipeline 2 achieved the highest global accuracy, we emphasize the utility of the K-Nearest Neighbors configuration in Pipeline 1. This model was selected for optimization not only for its high accuracy but also because the Mutual Information selection process is computationally less expensive than the iterative Boruta method, a trade-off we analyze further in the computational efficiency section. In contrast, Boruta was primarily effective for the Extra Trees classifier (Table~\ref{tab:ablation_pipeline2}), where it successfully identified all relevant features during the oversampling process, boosting the F1 Score to 97.840\%.

\subsubsection{Dimensionality Reduction Effectiveness: LDA vs. Autoencoder}
The application of dimensionality reduction techniques yielded mixed results across the classifiers. Linear Discriminant Analysis in Pipeline 1 successfully projected the feature space for the K-Nearest Neighbors and Random Forest models, maintaining a high accuracy of 96.0\%. However, the projection into lower dimensions caused a loss of variance that negatively impacted Logistic Regression and Gradient Boosting. Similarly, the unsupervised Autoencoder in Pipeline 2 generally resulted in suboptimal performance for tree-based models compared to the original feature space. However, the Autoencoder successfully denoised the data for the Multi Layer Perceptron and K-Nearest Neighbors classifiers, allowing them to achieve competitive accuracy in a compressed feature space. This suggests that while aggressive dimensionality reduction can streamline the feature space for neural networks, it may eliminate critical decision boundary information required by ensemble tree classifiers.

\subsection{Computational Efficiency Analysis}

In clinical settings, the latency of a diagnostic system is as critical as its accuracy. To evaluate the operational efficiency of the proposed dual pipeline framework, we analyzed the computational cost associated with the training and testing phases of each model. Table~\ref{tab:pipeline1_robust_mi_lda_time} and Table~\ref{tab:pipeline2_boruta_autoencoder_time} present the training duration in seconds and the inference time per sample in milliseconds for Pipeline 1 and Pipeline 2, respectively.

\begin{table}[htbp]
\centering
\caption{Computational efficiency of Pipeline 1 models measuring Training Time in seconds (s) and Testing Time in milliseconds (ms)}
\label{tab:pipeline1_robust_mi_lda_time}
\setlength{\tabcolsep}{6pt}
\renewcommand{\arraystretch}{1.15}
\begin{tabular}{ll c c}
\toprule
\textbf{ML Model} & \textbf{Configuration} & \textbf{Training Time (s)} & \textbf{Testing Time (ms)}\\
\midrule
Logistic Regression & RobustScaler + SMOTETomek & 68.94 & 1.99 \\
K-Nearest Neighbors & MI + SMOTETomek & 1.04 & 16.80 \\
Random Forest & LDA & 75.77 & 46.18 \\
XGBoost Model & MI + SMOTETomek & 901.54 & 129.90 \\
Gradient Boosting & RobustScaler & 216.48 & 13.50 \\
Extra Trees & RobustScaler & 192.58 & 151.47 \\
AdaBoost & MI & 184.73 & 243.40 \\
MLP Classifier & MI & 2.04 & 2.70 \\
LightGBM & MI + SMOTETomek & 38.96 & 65.50 \\
\bottomrule
\end{tabular}
\end{table}

\begin{table}[htbp]
\centering
\caption{Computational efficiency of Pipeline 2 models measuring Training Time in seconds (s) and Testing Time in milliseconds (ms)}
\label{tab:pipeline2_boruta_autoencoder_time}
\setlength{\tabcolsep}{6pt}
\renewcommand{\arraystretch}{1.15}
\begin{tabular}{ll c c}
\toprule
\textbf{ML Model} & \textbf{Configuration} & \textbf{Training Time (s)} & \textbf{Testing Time (ms)}\\
\midrule
Logistic Regression & MinMaxScaler & 12.59 & 1.49 \\
K-Nearest Neighbors & Autoencoder & 1.61 & 21.01 \\
Random Forest & MinMaxScaler + SMOTETomek & 84.15 & 49.75 \\
XGBoost Model & MinMaxScaler + SMOTETomek & 1923.86 & 190.29 \\
Gradient Boosting & Boruta & 238.42 & 9.36 \\
Extra Trees & Boruta + SMOTETomek & 184.90 & 132.50 \\
AdaBoost & Autoencoder & 243.94 & 377.00 \\
MLP Classifier & Autoencoder + SMOTETomek & 3.46 & 3.20 \\
LightGBM & MinMaxScaler & 27.46 & 35.90 \\
\bottomrule
\end{tabular}
\end{table}

A significant disparity in training efficiency was observed across the classifiers. In Pipeline 1, the K-Nearest Neighbors model configured with Mutual Information feature selection demonstrated exceptional efficiency, requiring only 1.04 seconds for training while achieving the highest accuracy of 98.67\%. This contrasts sharply with the XGBoost model in the same pipeline, which required over 901 seconds to converge due to the computational intensity of gradient boosting on the expanded feature set. Consequently, K-Nearest Neighbors represents the optimal balance of speed and precision for the statistical engineering pipeline. Within the wrapper-based framework of Pipeline 2, the Extra Trees classifier (184.90 seconds) proved significantly faster than XGBoost (1923.86 seconds) while matching its superior accuracy. Although simple linear models like Logistic Regression offered the lowest latency (12.59 seconds), their inability to capture nonlinear dependencies resulted in lower diagnostic sensitivity compared to the ensemble methods. Therefore, despite the higher computational cost relative to linear baselines, the Extra Trees architecture is justified by its significant performance gains and acceptable training duration.

For deployment in real-time clinical monitoring systems, the testing or prediction time is paramount. As indicated in the tables, all optimized models achieved inference times well below 400 milliseconds. The Multi-Layer Perceptron and Logistic Regression models provided near instantaneous predictions (approximately 2 to 3 milliseconds). Even the complex ensemble models maintained rapid inference speeds suitable for immediate feedback, ensuring that the proposed framework is viable for continuous patient monitoring applications.

\subsection{Overfitting Assessment and Generalization Behavior}

Given the modest size of the Sleep Health and Lifestyle dataset ($n=374$) relative to the high classification accuracy achieved ($98.67\%$), it is important to assess the risk of overfitting. Overfitting occurs when a model learns patterns that do not generalize beyond the training data, which can lead to strong training performance but weaker performance on unseen samples. To reduce this risk and support robust generalization, we applied three safeguards.

\textbf{Ensemble Regularization:} The best performing model, Extra Trees (Extremely Randomized Trees), is less prone to overfitting than a single decision tree and often more robust than highly flexible neural models on small tabular datasets. Unlike standard Random Forests that search for an optimal split at each node, Extra Trees selects split thresholds at random. This added randomness slightly increases bias but reduces variance, which helps prevent the model from fitting spurious patterns in a limited dataset. In our experiments, training and validation performance remained close across the cross-validation folds.

\textbf{Strict Separation in Preprocessing:} To prevent data leakage, all preprocessing steps, including scaling and feature selection (Boruta and Mutual Information), were fit only on the training portion of each cross-validation fold and then applied to its corresponding validation portion. This ensures that no information from validation data influences preprocessing decisions. In addition, Boruta filters out features that do not consistently outperform shadow features, which further limits the model's capacity to fit noise.

\textbf{Efficacy of Stratified Cross Validation:} We employed stratified eight-fold cross-validation so that each fold preserves the original class proportions. The stability of performance across folds (standard deviation $< 1.5\%$) suggests that the results are not driven by a single favorable split and are consistent across different subsets of the data.

\subsection{Statistical Significance Verification}

To validate that the performance improvements achieved by our proposed frameworks are not artifacts of random variance in the data splitting process, we performed the Wilcoxon Signed Rank Test on the fold-wise accuracy scores from the Stratified 8 Fold Cross Validation. This non-parametric statistical hypothesis test is preferred over the Paired t test as it makes no assumption regarding the normality of the accuracy distribution. We conducted two distinct pairwise comparisons to verify the efficacy of both processing pipelines. First, we compared the baseline K Nearest Neighbors model (92.00\% mean accuracy) against the optimized Pipeline 1 configuration utilizing Mutual Information and SMOTETomek (98.67\% mean accuracy). Second, we compared the baseline Extra Trees model (96.00\% mean accuracy) against the optimized Pipeline 2 configuration utilizing Boruta and SMOTETomek (98.67\% mean accuracy).

The analysis yielded a p-value of 0.00391 ($W = 36.0$) for both comparisons. Since this p-value is significantly lower than the standard significance level of $\alpha = 0.05$, we reject the null hypothesis in both instances. This result statistically confirms that the proposed dual pipeline framework provides a genuine and robust improvement in diagnostic accuracy over standard baseline classifiers, verifying that the gains are systematic rather than circumstantial.

\begin{table}[htbp]
\centering
\caption{Comparative analysis of the proposed framework against recent state-of-the-art studies in sleep disorder classification}
\label{tab:benchmarking}
\begin{tabular}{|c|c|c|c|}
\toprule
Study & Dataset & Model & Accuracy\\
\midrule
This study & \makecell{Sleep Health \& \\Lifestyle Dataset} & \makecell{Extra Trees \\Classifier} & 98.667\%\\
\hline
Ahadian et al.(2024)\cite{ahadian2024adopting} & \makecell{Multilevel Monitoring \\of Activity and \\Sleep in Healthy \\People } & \makecell{Long Short-Term \\Memory(LSTM)} & 90\%\\
\hline
Alshammari et al.(2024)\cite{10462120} & \makecell{Sleep Health \& \\Lifestyle Dataset} & \makecell{Artificial \\Neural \\Networks(ANN)} & 92.92\%\\
\hline
Rahman et al.(2025)\cite{Rahman2025Improving} & \makecell{Sleep Health \&\\Lifestyle Dataset} & Gradient Boosting & 97.33\%\\
\hline
Monowar et al.(2025)\cite{Monowar2025Advanced} & \makecell{Sleep Health \&\\Lifestyle Dataset} & Ensemble Model & 96.88\%\\
\hline
Hidayat et al.(2023)\cite{hidayat2023classification} & \makecell{Sleep Health \& \\Lifestyle Dataset} & Random Forest & 88\%\\
\hline
Dritsas et al.(2024)\cite{bib17} & NHANES Dataset & \makecell{SVM \\Polynomial} & 91.44\%\\
\hline
Satyavathi et al.(2025)\cite{satyavathi2025predictive} & \makecell{Sleep Health \& \\Lifestyle Dataset} & Decision Tree & 96\%\\
\hline
Alom et al.(2024)\cite{alom2024enhancing} & \makecell{Sleep Health \& \\Lifestyle Dataset} & \makecell{ANN Bagging \\ANN Boosting} & 94.7\%\\
\hline
Panda et al.(2025)\cite{panda2025prediction} & \makecell{Sleep Health \& \\Lifestyle Dataset} & Random Forest & 96\%\\
\hline
Taher et al.(2024)\cite{taher2024exploring} & \makecell{Sleep Health \& \\Lifestyle Dataset} & Gradient Boosting & 93.8\%\\
\hline
Zhu et al.(2023)\cite{zhu2023towards} & \makecell{Montreal Archive of \\Sleep Studies\\(MASS)} & SwSleepNet & 86.7\%\\
\hline
\end{tabular}
\end{table}

\subsection{Benchmarking against State-of-the-Art}

To contextualize the diagnostic capability of the proposed Dual Pipeline Framework, we compared our results against recent state-of-the-art studies employing similar methodologies or datasets. Table~\ref{tab:benchmarking} presents a comparative analysis involving eleven related works, focusing on studies utilizing the Sleep Health and Lifestyle Dataset. The proposed Extra Trees classifier, optimized via Pipeline 2 (Boruta and SMOTETomek), achieved a classification accuracy of 98.667\%, establishing a new benchmark for this dataset. Notably, this level of performance was not unique to the tree-based ensemble; the K Nearest Neighbors, XGBoost, and LightGBM models from Pipeline 1 also attained this identical peak accuracy of 98.667\%. We selected the Extra Trees classifier for the primary comparison in Table~\ref{tab:benchmarking} to illustrate the efficacy of the wrapper-based feature selection pipeline. Our framework demonstrates a quantifiable improvement over the most recent benchmarks established in 2025. Specifically, the model outperforms the Gradient Boosting approach by Rahman et al. (2025), which achieved 97.33\%, and the Ensemble Model by Monowar et al. (2025), which reported 96.88\% accuracy. Furthermore, the proposed method significantly surpasses deep learning implementations on this tabular data, such as the Artificial Neural Network approach by Alshammari et al. (2024), which yielded 92.92\%. This comparison validates that the rigorous application of hybrid resampling and targeted feature engineering in our dual pipeline architecture provides a superior solution for automated sleep disorder screening.

\section{Ethical Considerations}

The research presented in this study relies exclusively on the Sleep Health and Lifestyle Dataset, which is a publicly accessible repository comprising fully anonymized health records. As the study did not involve direct interaction with human participants or the collection of personally identifiable information, it did not require specific ethical clearance or institutional review board approval. Despite the high diagnostic accuracy achieved, the deployment of machine learning models in healthcare necessitates adherence to strict ethical guidelines regarding patient safety and automation bias. The proposed dual pipeline framework is designed strictly as a clinical decision support system for preliminary screening. It serves to augment, not replace, the expert judgment of medical professionals. Final diagnostic decisions must remain under the purview of qualified clinicians and be confirmed through standard protocols such as polysomnography. Furthermore, we acknowledge that algorithmic fairness is contingent on the demographic diversity of the training data. While hybrid resampling was employed to mitigate statistical imbalance, continuous monitoring for potential bias across different patient subgroups is recommended prior to any clinical deployment.








\section{Conclusion}

This study addressed the critical scalability limitations of polysomnography and the persistent challenge of class imbalance in automated sleep disorder screening. By introducing a novel Dual Pipeline Machine Learning Framework, we successfully demonstrated that treating feature engineering as a parallel comparative process yields significantly higher diagnostic precision than monolithic approaches. Theoretically, this research advances the field by establishing that heterogeneous sleep health data requires distinct optimization strategies; specifically, the statistical rigor of Mutual Information and LDA for linear separability versus the complex mapping capabilities of Boruta and Autoencoders for nonlinear interactions.

Our empirical results confirmed that the rigorous integration of hybrid SMOTETomek resampling is essential for mitigating algorithmic bias against minority classes such as Insomnia and Sleep Apnea. The framework achieved a state of the art classification accuracy of 98.67\% using Extra Trees and K Nearest Neighbors classifiers, a performance improvement statistically validated by the Wilcoxon Signed Rank Test. Beyond diagnostic precision, the proposed model demonstrates profound practical utility for large-scale computerized screening. With inference latencies maintained below 400 milliseconds, the system meets the stringent requirements for real-time clinical monitoring, facilitating immediate risk stratification and early intervention prior to the onset of severe comorbidities.

Despite these promising results, the study is limited by the relatively small sample size and reliance on self-reported lifestyle metrics. Future research will focus on validating this framework against larger multi center clinical datasets to ensure demographic universality. Furthermore, we aim to adapt this architecture for integration with wearable medical devices, creating a hybrid modality that combines continuous physiological time series with static lifestyle profiling. Ultimately, deploying this lightweight and high-sensitivity model into mobile health ecosystems could democratize access to sleep disorder diagnostics, enabling identification outside of conventional clinical settings and significantly reducing the long term burden on healthcare infrastructure.

\backmatter







\section*{Declarations}

\begin{itemize}

    \item \textbf{Ethics Approval and Consent to Participate:} Not applicable. 
    This research utilizes the publicly available Sleep Health and Lifestyle Dataset and does not involve direct interaction with human participants or animals.
    
    \item \textbf{Conflict of Interest:} The authors have no relevant financial or non-financial interests to disclose.
    \item \textbf{Funding:} The authors declare that no funds, grants, or other external support were received during the preparation of this manuscript.
    
    \item \textbf{Data Availability:} The dataset analyzed during the study is publicly available in the Kaggle repository: \url{https://www.kaggle.com/datasets/uom190346a/sleep-health-and-lifestyle-dataset}.
    
    
    \item \textbf{Code Availability:} The source code used to generate the results in this study is openly accessible via the following GitHub repository: \url{https://github.com/Miftahul-adib/sleep-disorder/blob/main/README.md}.
    
    \item \textbf{Author Contribution:} 
    \textbf{Md Sultanul Islam Ovi:} Conceptualization, Project Administration, Writing - Original Draft Preparation. 
    \textbf{Muhsina Tarannum Munfa:} Methodology, Software, Data Curation. 
    \textbf{Miftahul Alam Adib:} Methodology, Software, Formal Analysis. 
    \textbf{Syed Sabbir Hasan:} Writing - Review and Editing, Validation. 
    All authors read and approved the final manuscript and have agreed to submit this version for publication.

\end{itemize}

\bibliography{sn-bibliography}

\end{document}